\documentclass[]{llncs}
\usepackage{graphicx}

\usepackage{tikz}
\usepackage{comment} 
\usepackage{amsmath,amssymb} 
\usepackage{color}
\usepackage{algorithm}
\usepackage{listings}
\usepackage{boldline}
\usepackage{subfigure}
\usepackage{makecell}

\begin{document}
\pagestyle{headings}
\mainmatter
\def\ECCVSubNumber{100}  

\title{Rethinking Token-Mixing MLP for MLP-based Vision Backbone} 

\titlerunning{Rethinking Token-Mixing MLP}
%
\author{Tan Yu, Xu Li, Yunfeng Cai, Mingming Sun,  Ping Li\\
}
\institute{
Cognitive Computing Lab\\
Baidu Research\\
10900 NE 8th St. Bellevue, Washington 98004, USA\\
No.10 Xibeiwang East Road, Beijing 100193, China\\
{\tt\small \{tanyu01,lixu13,caiyunfeng,sunmingming01,liping11\}@baidu.com}
}
\maketitle

\begin{abstract}
In the past decade, we have witnessed rapid progress in the machine vision backbone.  By introducing the inductive bias from the image processing, convolution neural network (CNN) has achieved excellent performance in numerous computer vision tasks and has been established as \emph{de facto}  backbone. In recent years,  inspired by the great success achieved by Transformer in NLP tasks, vision Transformer models emerge. Using much less inductive bias, they have achieved promising performance in computer vision tasks compared with their CNN counterparts. More recently, researchers investigate  using the pure-MLP architecture to build the vision backbone to further reduce the inductive bias,  achieving good performance.  The pure-MLP backbone is built upon channel-mixing MLPs to fuse the channels and token-mixing MLPs for communications between patches.  In this paper, we re-think the design of the token-mixing MLP.  We discover that token-mixing MLPs in existing MLP-based backbones are spatial-specific, and thus it is sensitive to spatial translation. Meanwhile, the channel-agnostic property of the existing token-mixing MLPs limits their capability in mixing tokens. To overcome those limitations, we propose an improved structure termed as Circulant Channel-Specific  (CCS) token-mixing MLP, which is spatial-invariant and channel-specific. It takes fewer parameters but achieves higher classification accuracy on ImageNet1K benchmark. 
\end{abstract}

\section{Introduction}
\label{sec:intro}
Convolution neural network (CNN)~\cite{krizhevsky2012imagenet,he2016deep} has been \emph{de facto} backbone for computer vision in the past years. Recently,  inspired by the accomplishment achieved by Transformer~\cite{vaswani2017attention} in NLP, several vision Transformer methods emerge~\cite{dosovitskiy2020image,touvron2020training}. Compared with CNN,  vision Transformer methods do not need hand-crafted convolution kernels and simply stack a few Transformer blocks. Albeit vision Transformer methods take much less inductive bias, they have achieved comparable or even better recognition accuracy than CNN models.  More recently, researchers have taken a step forward. They propose MLP-based models~\cite{tolstikhin2021mlp,touvron2021resmlp,melas2021you}  with only MLP layers taking even less inductive bias.

MLP-mixer~\cite{tolstikhin2021mlp} is the pioneering work of pure-MLP  vision backbone.  
It is built on two types of MLP layers: the channel-mixing MLP layer and the token-mixing MLP layer.  The channel-mixing MLP is for communications between the channels. In parallel, the token-mixing MLP conducts communications between patches.  Compared with vision Transformer adaptively assigning attention based on the relations between patches,   the token-mixing MLP assigns fixed attention to patches based on their spatial locations.  In fact, the channel-mixing MLP is closely related with depthwise convolution~\cite{chollet2017xception,howard2017mobilenets,kaiser2018depthwise}. The differences between token-mixing MLP and depthwise convolution are three-fold. Firstly, the token-mixing MLP has a global reception field but the depthwise convolution has only a local reception field.  The global reception field enables the token-mixer MLP to have access to the whole visual content in the image.  Secondly, the depthwise convolution is spatial-invariant, whereas the channel-mixing MLP  no longer possesses the spatial-invariant property and thus its output is sensitive to spatial translation. Lastly, for a specific position, the token-mixing MLP assigns an identical weight to elements in different channels.  In contrast, the depthwise convolution applies different convolution kernels on different channels for encoding richer visual patterns.

Observing the strength and weakness of the vanilla token-mixing MLP in existing MLP-based backbones, we propose an improved structure, the Circulant Channel-Specific (CCS) token-mixing MLP which preserves the strength of the existing token-mixing MLP and overcomes its weakness. Similar to the vanilla token-mixing MLP, our CCS token-mixing MLP has a global reception field. Meanwhile, we adopt a circulant structure in the weight matrix to achieve the spatial-invariant property. Moreover, we adopt a channel-specific design to exploit richer manners in mixing tokens.  In Table~\ref{tab:prop}, we compare the properties of our CCS with depthwise convolution and vanilla token-mixing MLP.
Moreover, benefited from circulant structure, our CCS needs considerably fewer parameters than the token-mixing MLP.
To be specific, CCS only needs $GN$ parameters where $N$ is the number of patches and $G$ is the number of groups. In contrast, the token-mixing MLP normally takes $N^2$ parameters, which are significantly more than CCS since $G \ll N$. By replacing the token-mixing MLP with the proposed CCS in two existing pure-MLP vision backbones, we achieve consistently higher recognition accuracy on ImageNet1K with fewer parameters.

\begin{table}
\centering
\begin{tabular}{c|c|c|c} \hlineB{3}
   Property          & Reception & Spatial     & Channel     \\
   \hline 
Depthwise    & local     & agnostic & specific    \\
Token-mixing & global    & specific   & agnostic  \\
CSC        & global    & agnostic & group-specific \\ \hlineB{3}
\end{tabular}
\vspace{0.1in}
\caption{The properties of depthwise convolution, the token-mixing MLP in MLP-mixer~\cite{tolstikhin2021mlp}  and our Circulant Channel-Specific (CCS) token-mixing MLP.}
\label{tab:prop}
\end{table}

\section{Related Work}

\subsection{Vision Transformer}

Vision Transformer (ViT)~\cite{dosovitskiy2020image}   is the pioneering work adopting the architecture solely with Transformer layers for computer vision tasks. It crops an image into non-overlap patches and feeds these patches through a stack of Transformer layers for attaining communications between patches.  Using less hand-crafted design, ViT achieves competitive recognition accuracy compared with its CNN counterparts. Nevertheless, it requires a huge scale of images for pre-training. DeiT~\cite{touvron2020training}  adopts a more advanced data augmentation method as well as a stronger optimizer, achieving excellent performance through training on a medium-scale dataset. PVT~\cite{wang2021pyramid} introduces a  progressive shrinking pyramid into ViT, improving the recognition accuracy.   PiT~\cite{heo2021pit} integrates depthwise convolution between Transformer blocks and also devises a  shrinking pyramid structure. In parallel, Tokens-to-Tokens ViT~\cite{yuan2021tokens} effectively models the local structure through recursively aggregating neighboring tokens. It achieves higher recognition accuracy with fewer FLOPs.  Transformer-iN-Transformer (TNT)~\cite{han2021transformer} also focuses on modeling the local structure. It devises an additional Transformer to model the intrinsic structure information inside each patch. Recently, the focus of vision Transformer is on improving efficiency through exploiting locality and sparsity.  For instance, Swin~\cite{liu2021swin} develops a hierarchical backbone with local-window Transformer layers with variable window sizes.  Exploiting local windows considerably improves the efficiency, and the variable window scopes achieve the global reception field. Twins~\cite{chu2021twins}  alternately stacks a local-dense Transformer and a global-sparse Transformer,  also achieving a global reception field in an efficient manner. Shuffle Transformer~\cite{huang2021shuffle} also adopts the local-window Transformer and achieves the cross-window connections through spatially shuffling.  S$^2$ViTE~\cite{chen2021chasing} explores on integrating sparsity in vision Transformer and improves the efficiency from both model and data perspectives. Recent works~\cite{gao2021container,wu2021cvt} investigate combining convolution and Transformer to build a hybrid vision backbone.

\subsection{MLP-based Backbone}
Recently, MLP-mixer~\cite{tolstikhin2021mlp} proposes a pure-MLP backbone with less inductive bias than vision Transformer and CNNs, achieving excellent performance in image recognition. It is built upon two types of MLP layers: the channel-mixing MLP and the token-mixing MLP. The channel-mixing MLP is equivalent to $1\times 1$ convolution layer. It achieves the communications between channels.  The token-mixing MLP   achieves cross-patch communications. It is similar to the self-attention block in Transformer. But the attention in Transformer is dependent on the input patches, whereas the attention in the token-mixing MLP is agnostic to the input.  Feed-forward (FF)~\cite{melas2021you} adopts a similar architecture as MLP-mixer, also achieving excellent performance.  ResMLP~\cite{touvron2021resmlp} simplifies the token-mixing MLP in MLP-mixer and adopts a deep architecture stacking a huge number of layers. Meanwhile, to stabilize the training, ResMLP proposes an affine transform layer to replace the layer normalization in MLP-mixer.  Benefited from exploiting a deeper architecture, ResMLP achieves a better performance than MLP-mixer. Meanwhile, by smartly trading off the hidden size and depth, ResMLP takes fewer FLOPs and fewer parameters than MLP-mixer. Recently, gMLP~\cite{liu2021pay} exploits a gating operating to enhance the effectiveness of the token-mixing MLP, attaining a higher recognition accuracy than MLP-mixer.  In parallel,  External Attention~\cite{guo2021beyond} replaces the self-attention operation with the attention on external memory implemented by fully-connected layers, achieving comparable performance with vision Transformers. Recently, spatial-shift MLP (S$^2$MLP)~\cite{yu2021s} adopts a spatial-shift operation~\cite{wu2018shift,brown20194} to achieve the communications between patches, achieving excellent performance.


\section{Preliminary}
In this section, we briefly review the existing mainstream  MLP backbone, MLP-Mixer~\cite{tolstikhin2021mlp}.
It consists of three modules including a per-patch fully-connected layer, a stack of $N$ Mixer layers, and a fully-connected layer for classification. 

\vspace{0.15in}

\noindent \textbf{Input.} For an input image of the size $W \times H \times 3$,  we  crop it into $N$ non-overlap patches of $p \times p \times 3$ size.
 For each patch, it is unfolded into a vector $\mathbf{p} \in \mathbb{R}^{3p^2}$. In total, we obtain a set of patch feature vectors $\mathcal{P} = \{\mathbf{p}_0,\cdots,\mathbf{p}_{N-1} \}$, which are the input of the MLP-Mixer. 

\vspace{0.2in}
\noindent \textbf{Per-patch fully-connected layer.}  For each patch feature $\mathbf{p}_i \in \mathcal{P}$,  the per-patch fully-connected layer maps it into a $C$-dimensional vector by
\begin{equation}
\mathbf{x}_i \gets \mathbf{W}_{0} \mathbf{p}_i + \mathbf{b}_0, 
\end{equation}
where $\mathbf{W}_0 \in \mathbb{R}^{3p^2\times C}$ and $\mathbf{b}_0 \in \mathbb{R}^{C}$ are the weights of the per-patch fully-connected layer.

\vspace{0.2in}
\noindent \textbf{Mixer layers.} MLP-mixer stacks $L$ mixer layers of the same size, and each layer contains two MLP blocks: the token-mixing MLP block and the channel-mixing MLP block. Let us denote the patch features in the input  of each mixer layer as $\mathbf{X} = [\mathbf{x}_0,\cdots,\mathbf{x}_{N-1}] \in \mathbb{R}^{C\times N}$ where $C$ is the number of channels and $N$ is the number of patches. Channel-mixing block projects patch features $\mathbf{X}$ along the channel dimension by
\begin{equation}
\mathbf{U} = \mathbf{U} + \mathbf{W}_2 \sigma[\mathbf{W}_1 \mathrm{LayerNorm}(\mathbf{X})].
\end{equation}
$\mathbf{W}_1 \in \mathbb{R}^{rC\times C}$ represents weights of a fully-connected layer increasing the feature dimension from $C$ to $rC$  where $r>1$ is the expansion ratio.
$\mathbf{W}_2 \in \mathbb{R}^{C\times rC}$ denotes weights of a fully-connected layer  decreasing the feature dimension from $rC$ back to $C$.
$\mathrm{LayerNorm}(\cdot)$ denotes the layer normalization~\cite{ba2016layer}  widely used in Transformer-based models and $\sigma(\cdot)$ denotes the activation function implemented by GELU~\cite{hendrycks2016gaussian}. Then the output of the channel-mixing block $\mathbf{U}$ is fed into the token-mixing block for communications between patches: 
\begin{equation}
\label{original}
\mathbf{Y} = \mathbf{U} + \sigma[ \mathrm{LayerNorm}(\mathbf{U})\mathbf{W}_3] \mathbf{W}_4,
\end{equation}
where $\mathbf{W}_3 \in \mathbb{R}^{N\times M}$ and $\mathbf{W}_4 \in \mathbb{R}^{M\times N}$  denote the weights of  fully-connected layers projecting patch features $\mathbf{U}$  along the token dimension. Recently, ResMLP~\cite{touvron2021resmlp}  adopts a simplified  token-mixing block:
\begin{equation}
\label{simple}
\mathbf{Y} = \mathbf{U} + \mathrm{LayerNorm}(\mathbf{U})\mathbf{W}_3,
\end{equation}
where $\mathbf{W}_3 \in \mathbb{R}^{N\times N}$ is a square matrix. The experiments in ResMLP~\cite{touvron2021resmlp} show that the simplified token-mixing block achieves comparable performance as the original one defined in Eq.~\eqref{original}. Unless otherwise specified, the token-mixing  we mention below is the simplified form defined in Eq.~\eqref{simple}.

\vspace{0.2in}
\noindent \textbf{Classification head.}  After a stack of $L$ mixer layers,  $N$ patches features are generated. They are further aggregated into a global vector through average pooling. Then the global vector is fed into a fully-connected layer for classification.

\section{Circulant  Channel-Specific Token-mixing MLP}
\begin{figure}[h!]
   \centering
    \subfigure[Vanilla token-mixing MLP.]{\includegraphics[scale=0.85]{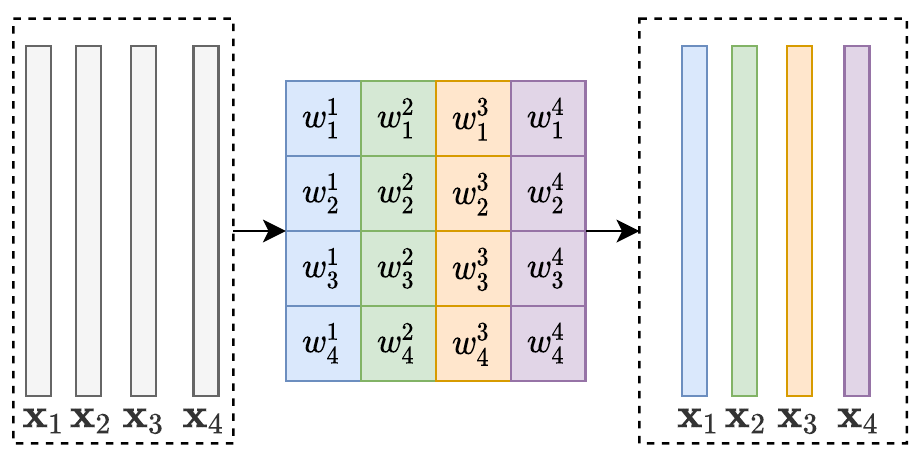}} 
    
    \vspace{0.1in}
    
    \subfigure[Our CCS.]{\includegraphics[scale=0.85]{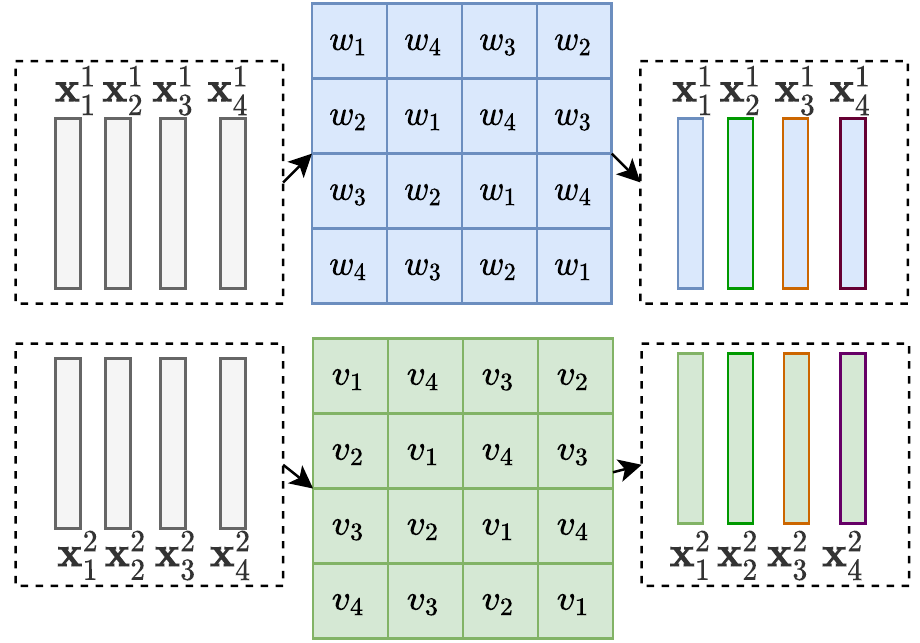}}
    
    \vspace{-0.1in}
    
    \caption{Comparisons between vanilla token-mixing MLP in existing MLP-based backbones and the proposed Circulant Channel-Specific (CCS) token-mixing MLP. The proposed CCS imposes a circulant-structure constraint on the weights of the token-mixing MLP. Meanwhile, our CCS splits the input $\mathbf{X}$ into multiple groups along the channel dimension. It uses different weight matrices on different groups for mixing tokens in a more flexible way.}
    \label{fig:cp}\vspace{-0.08in}
\end{figure}

In this section, we introduce our circulant channel-specific (CCS) token-mixing MLP. As we mention in the introduction, the vanilla token-mixing MLPs in  MLP-mixer and ResMLP are spatial-specific and channel-agnostic.
The spatial-specific property makes the token-mixing MLP sensitive to spatial translation. Meanwhile, the channel-agnostic configuration limits its capability in mixing tokens. The motivation of designing CCS token-mixing MLP is to achieve the spatial-agnostic property and meanwhile attain a channel-specific configuration for mixing tokens in a more effective way. To this end, we make two modifications to the vanilla token-mixing MLP, as illustrated in Figure~\ref{fig:cp}, Algorithm~\ref{alg:code}, and details below.

\subsection{Circulant structure}
 We devise the weight matrix $\mathbf{W}_3$ in Eq.~\eqref{simple} in a circulant structure.  To be specific, we set the weight matrix $\mathbf{W}_3$ in the below form:
\begin{equation}
\label{kro}
\mathbf{W}_3 =  \begin{bmatrix}
w_0 & w_{N-1} & w_{N-2} & \dots & w_2 & w_{1}\\
w_1 & w_0& w_{N-1} &\dots & w_3 &w_{2} \\
w_2 & w_1& w_0 & \dots & w_4 & w_{3} \\
\vdots & \vdots & \vdots & \ddots &\vdots & \vdots \\
w_{N-2} & w_{N-3} & w_{N-4} & \dots & w_{0} & w_{N-1} \\
w_{N-1} &  w_{N-2} & w_{N-3} &\dots & w_1 & w_{0}
\end{bmatrix}.
\end{equation}
It is  fully specified by one vector, $\mathbf{w} = [w_0,\cdots,w_{N-1}]$, which appears as the first column of $\mathbf{W}_3$.  The circulant structure is naturally spatial-agnostic for mixing tokens. Below we show it in detail. 
Let us denote $\mathrm{LayerNorm}(\mathbf{U})$ in Eq.~\eqref{simple} by $\hat{\mathbf{U}} = [\hat{\mathbf{u}}_0, \cdots, \hat{\mathbf{u}}_{N-1}]$, where each item $\hat{\mathbf{u}}_i$ ($i\in[0,N-1]$) is the patch feature after layer normalization.  We rewrite Eq.~\eqref{simple} into
\begin{align}
\mathbf{Y} = \mathbf{U} + \hat{\mathbf{U}} \mathbf{W}_3.
\end{align}
Let us denote the $i$-th column of $\hat{\mathbf{U}}$ by $\hat{\mathbf{u}}_i$ and  the $i$-th column of $\hat{\mathbf{U}}\mathbf{W}_3$ by $\bar{\mathbf{u}}_i$. Then it is straightforward to obtain that
\begin{equation}
\label{invariant}
\bar{\mathbf{u}}_i =  \sum_{j=0}^{N-1} {w}_{j} \hat{\mathbf{u}}_{(i+j) \% N }, \forall i \in [0,N-1],
\end{equation}
where $\%$ denotes the modulo operation.
As shown in  Eq.~\eqref{invariant}, the mixing operation for obtaining each patch feature $\bar{\mathbf{u}}_i$ is invariant to the location $i$.  Thus, the token-mixing MLP with the circulant-structure $\mathbf{W}_3$ is spatial-agnostic, which is not sensitive to spatial translation.  Meanwhile, the circulant structure reduces the number of parameters from $N^2$ to $N$. At the same time, the multiplications between a $N$-dimension vector and an $N\times N$ circulant matrix only take $\mathcal{O}(N\mathrm{log} N)$ complexity using fast Fourier transform (FFT), which is more efficient than the vanilla vector-matrix multiplication with $\mathcal{O}(N^2)$ computational complexity. Specifically, given an vector $\mathbf{x} \in \mathbb{R}^N$ and a circulant matrix $\mathbf{W}$ with the first column $\mathbf{w}$, the matrix-vector multiplication $\mathbf{W}\mathbf{x}$ can be computed efficiently through

\begin{equation}
\mathbf{x}\mathbf{W} = \mathrm{FFT}[ \mathrm{IFFT}(\mathbf{x}) \odot   \mathrm{FFT}(\mathbf{w})], 
\end{equation}

\noindent where FFT denotes fast Fourier transform, IFFT denotes inverse fast Fourier transform, and $\odot$ denotes the element-wise product. Since both FFT and IFFT take only $\mathcal{O}(N\mathrm{log}N)$ computational complexity, the total complexity is only $\mathcal{O}(N\mathrm{log}N)$. In contrast,  the vanilla token-mixing MLP in existing methods takes $\mathcal{O}(N^2)$ computational complexity. 
But when the number of patches, $N$, is not large, FFT with  $\mathcal{O}(N\mathrm{log} N)$  complexity can not demonstrate its efficiency advantage over the vanilla token-mixing MLP with $\mathcal{O}(N^2)$  complexity.

\begin{algorithm}[t]
\caption{Pseudocode of our Circulant Channel-Specific (CCS) token-mixing MLP.}
\label{alg:code}
\lstset{
  backgroundcolor=\color{white},
  basicstyle=\fontsize{9.5pt}{9.5pt}\ttfamily\selectfont,
  columns=fullflexible,
  breaklines=true,
  captionpos=b,
  commentstyle=\fontsize{9.5pt}{9.5pt}\color{codeblue},
  keywordstyle=\fontsize{9.5pt}{9.5pt},
}
\begin{lstlisting}[language=Python]
class CCS(nn.Module):
    def __init__(self, groups, patches):
        super().__init__()
        self.groups = groups
        self.w = nn.Linear(patches,self.groups).weight
    def forward(self, x):
        B, N, C = x.shape
        x = ifft(x,dim=1)
        w = fft(self.w,dim=1).unsqueeze(0).unsqueeze(2).expand(B,N,C//self.groups,self.groups).reshape(B,N,C)
        x = x.mul(w)
        x = fft(x,dim=1).real
        return x
\end{lstlisting}
\end{algorithm}

\subsection{Channel-Specific settings} The token-mixing MLP in MLP-mixer~\cite{tolstikhin2021mlp} shares the same weight among different channels.  That is, the same token-mixing MLP is applied to each of $C$ channels in $\hat{\mathbf{U}} \in \mathbb{R}^{C\times N}$. A straightforward extension is to devise $C$ separable MLPs, namely, $\{\mathrm{MLP}_c\}_{c=1}^C$. For each slide of $\hat{\mathbf{U}}$ along the channel, $\mathbf{U}[c,:] \in \mathbf{R}^N$, it is processed by a specific $\mathrm{MLP}_c$. This straightforward extension increases the number of parameters in the vanilla token-mixing MLP of MLP-mixer from $N^2$ to $N^2C$.  But the number of parameters, $N^2C$, is extremely huge,  making the network prone to over-fitting. Thus, it does not bring noticeable improvements in the recognition accuracy, as claimed in the supplementary material of MLP-mixer~\cite{tolstikhin2021mlp}.

 In contrast, in our circulant-structure settings, even if we adopt $C$ separable   MLPs to process different channels of $\hat{\mathbf{U}}$, the number of parameters only increases from $N$ to $NC$. The number of parameters is still on a medium scale.  Thus, as shown in our experiments,  using  $C$ separable circulant-structure  MLPs can achieve better performance than that with a single circulant-structure  MLP. To further reduce the number of parameters, we split $C$ channels into $G$ groups and each group contains $C/G$ channels.  For each group, we use a specific MLP. Thus, in this configuration, the number of parameters of our circulant-structure token-mixing MLP increases to $GN$, which is still significantly fewer than  $N^2$ parameters in vanilla MLP-mixing MLP since $G\ll N$. It is worth noting that, the computational complexity of our CCS token-mixing MLP is irrelevant to the number of groups, $G$. Thus, using $G$ MLPs in token-mixing does not bring more computational cost than using a single MLP.  We visualize the architecture of the vanilla token-mixing MLP  and our CCS token-mixing MLP in Figure~\ref{fig:cp}. We compare our CCS with the vanilla token-mixing MLP in Table~\ref{tab:cppr}.  As shown in the table, our CCS token-mixing MLP takes fewer parameters than the vanilla token-mixing.   We give the pseudocode of the proposed CCS token-mixing MLP in Algorithm~\ref{alg:code}.

\begin{table}
\centering
\begin{tabular}{ccc} \hlineB{3}
    settings         & $\#$ of parameters & \makecell{computational \\ complexity} \\ \hline

token-mixing & $N^2$   & $N^2C$   \\
CCS (ours)        & $GN$    & $N\mathrm{log}NC$  \\ \hlineB{3}
\end{tabular}
\vspace{0.1in}
\caption{Comparisons between our CCS and the vanilla token-mixing MLP on the number of parameters and computational complexity. $N$ is the number of patches, $C$ is the number of channels, and  $G$ denotes the number of groups, where $G\ll N$.}
\label{tab:cppr}\vspace{-0.15in}
\end{table}

\section{Experiments}
\textbf{Datasets.} The main results are  on ImageNet1K dataset~\cite{deng2009imagenet}. It contains $1.2$ million  images from one thousand categories for training and $50$ thousand  images for validation.  Note that MLP-mixer and ResMLP also exploit larger-scale datasets including  ImageNet21K~\cite{deng2009imagenet} and JFT-300M~\cite{sun2017revisiting} datasets. Nevertheless, due to limited computing resources, it is impractical for us  to train our models on these two datasets.

\begin{table}[htp!]
\centering
\begin{tabular}{cccccccc}
\hlineB{3}
default settings  & $N$ &$L$ & $C$ & $r$ & $p$ & $G$ & parameters  \\\hline
Mixer-B/16~\cite{tolstikhin2021mlp} + CCS  & $196$  & $12$ & $768$  & $4$   & $16$ & $8$ & $57$M  \\
ResMLP-36~\cite{touvron2021resmlp} + CCS   & $196$  & $36$ & $384$  & $4$   & $16$ &$8$ & $43$M \\
  \hlineB{3}
\end{tabular}
\vspace{0.1in}
\caption{ $N$ is the number of tokens,  $L$  is the number of blocks,  $C$ is the hidden size, $r$ is the expansion ratio,  $p$ is the patch size, and $G$ is the number of groups in Table~\ref{tab:cppr}.}
\label{tab:para}
\end{table}

\vspace{0.1in}
\noindent \textbf{Settings.} Our training follows the settings in DeiT~\cite{touvron2020training}. To be specific, we adopt AdamW~\cite{loshchilov2018decoupled} as the optimizer with weight decay 0.05. A linear warm-up process is conducted at the beginning of training.  The  learning rate is intially 1e-3 and gradually drops to 1e-5  in a cosine-decay manner. We utilize multiple data augmentation approaches including  random crop, random clip, Rand-Augment~\cite{cubuk2020randaugment}, Mixup~\cite{zhang2017mixup} and CutMix~\cite{yun2019cutmix}. Besides, we use replace dropout with DropPath~\cite{larsson2016fractalnet} and conduct label smoothing~\cite{szegedy2016rethinking} and repeated augmentation~\cite{hoffer2020augment}. The whole training process takes $300$ epochs.   We integrate the proposed CCS token-mixing MLP in two types of mainstream MLP-based backbones including Mixer-B/16~\cite{tolstikhin2021mlp}  and ResMLP-36~\cite{touvron2021resmlp}. Specifically, we only replace the vanilla token-mixing MLP with the proposed CCS token-mixing MLP and keep other layers unchanged. Table~\ref{tab:para} provides the detailed settings for these two backbones. The code will be available based on the PaddlePaddle deep learning platform.

\subsection{Main experimental results}
The main experimental results are presented in Table~\ref{main}. 
We compare ours with existing CNN-based models, Transformer-based models and MLP-based models. The first part of Table~\ref{main} contains the CNN-based models, including the classic ResNet-50~\cite{he2016deep}  and the recent state-of-the-art methods such as  RegNetY-16GF~\cite{radosavovic2020designing}, EfficientNet-B3~\cite{tan2019efficientnet} and EfficientNet-B5~\cite{tan2019efficientnet}. Note that,   the architectures of RegNetY-16GF, EfficientNet-B3 and EfficientNet-B5 are all attained based on  network architecture search (NAS).  The main strength of EfficientNet is  excellent recognition accuracy with the extremely cheap computational cost.  

\begin{table*}[htp!]
\centering
\begin{tabular}{c|c|c|c|c|c}
\hlineB{3}
Model & Resolution & Top-1 ($\%$) & Top5 ($\%$)  & Params (M) & FLOPs (B) \\ 
\hlineB{3}
\multicolumn{6}{c}{CNN-based methods}  \\ \hlineB{3}
ResNet50~\cite{he2016deep}  &  $224 \times 224$    &  $76.2$    &     $92.9$   &    $25.6$ & $4.1$   \\
RegNetY-16GF~\cite{radosavovic2020designing}  &  $224 \times 224$    &  $80.4$    &     $-$   &    $83.6$ & $15.9$   \\
EfficientNet-B3~\cite{tan2019efficientnet}  &  $300 \times 300$    &  $81.6$    &     $95.7$   &    $12$ & $1.8$  
\\
EfficientNet-B5~\cite{tan2019efficientnet}  &  $456 \times 456$    &  $84.0$    &     $96.8$   &    $30$ & $9.9$  
\\
\hlineB{3}
\multicolumn{6}{c}{Transformer-based methods}  \\ \hlineB{3}
ViT-B/16$^{*}$~\cite{dosovitskiy2020image,tolstikhin2021mlp}&  $224 \times 224$    &  $79.7$     & $-$       &    $86.4$ & $17.6$  \\
DeiT-B/16~\cite{touvron2020training} &   $224 \times 224$   &  $81.8$    & $-$       &  $86.4$ & $17.6$     \\
PVT-L~\cite{wang2021pyramid} &    $224 \times 224$   &  $82.3$    & $-$       &  $61.4$ & $9.8$     \\ 
TNT-B~\cite{han2021transformer} &    $224 \times 224$   &  $82.8$    & $96.3$       &  $65.6$ & $14.1$     \\
T2T-24~\cite{yuan2021tokens} &    $224 \times 224$   &  $82.6$    & $-$       &  $65.1$ & $15.0$     \\
CPVT-B~\cite{chu2021conditional} &    $224 \times 224$   &  $82.3$    & $-$       &  $88$ & $17.6$     \\
PiT-B/16~\cite{heo2021pit} &   $224 \times 224$   &  $82.0$    & $-$       &  $73.8$ & $12.5$     \\
CaiT-S32~\cite{touvron2021going} &    $224 \times 224$   &  $83.3$    & $-$       &  $68$ & $13.9$     \\
Swin-B~\cite{liu2021swin} &    $224 \times 224$   &  $83.3$    & $-$       &  $88$ & $15.4$    
\\
Nest-B~\cite{zhang2021aggregating} &    $224 \times 224$   &  $83.8$    & $-$       &  $68$ & $17.9$    \\
Container~\cite{gao2021container} &    $224 \times 224$   &  $82.7$    & $-$       &  $22.1$ & $8.1$    
\\ \hlineB{3}
\multicolumn{6}{c}{MLP-based methods}  \\ \hlineB{3}  
    FF~\cite{melas2021you} &    $224 \times 224$   &  $74.9$    & $-$       &  $59$ & $12$     \\ \hline
     S$^2$-MLP-wide~\cite{yu2021s} &    $224 \times 224$   &  $80.0$    & $94.8$       &  $71$ & $14$     \\  
    Mixer-B/16~\cite{tolstikhin2021mlp} &    $224 \times 224$   &  $76.4$    & $-$       &  $59$ & $12$     \\
    Mixer-B/16$^*$ &    $224 \times 224$   &  $77.2$    & $92.9$       &  $59$ & $12$     \\ 
     Mixer-B/16$^*$+CCS &    $224 \times 224$   &  $79.8$    & $94.6$       &  $57$ & $11$     \\\hline 
     S$^2$-MLP-deep~\cite{yu2021s} &    $224 \times 224$   &  $80.7$    & $95.4$       &  $51$ & $11$  \\
     ResMLP-36~\cite{touvron2021resmlp} &    $224 \times 224$   &  $79.7$    & $-$       &  $44$ & $9$     \\
     ResMLP-36$^*$&    $224 \times 224$   &  $79.8$    & $94.7$       &  $44$ & $9$     \\   
       ResMLP-36$^*$+CCS&    $224 \times 224$   &  $80.6$    & $95.3$       &  $43$ & $9$     \\   
        \hlineB{3}
\end{tabular}
\vspace{0.2in}
\caption{Comparisons with other models on ImageNet-1K benchmark without extra data.  ViT-B/16$^{*}$ denotes the result of ViT-B/16 model reported in MLP-Mixer~\cite{tolstikhin2021mlp} with extra regularization. Mixer-B/16$^*$ and ResMLP-36$^*$ denotes the results of our implementation of  Mixer-B/16~\cite{tolstikhin2021mlp} and  ResMLP-36~\cite{touvron2021resmlp} based on settings in DeiT~\cite{tolstikhin2021mlp}. ``+CCS'' denotes replacing the vanilla token-mixing MLP with the proposed CCS  token-mixing MLP.}
\label{main}
\end{table*}

The second group of methods is based on vision Transformer. ViT~\cite{dosovitskiy2020image} is the pioneering work with architecture solely based on Transformer. We show its performance with extra regularization reported in MLP-mixer~\cite{tolstikhin2021mlp}, which is better than the performance in the original ViT paper.  DeiT~\cite{touvron2020training} adopts more advanced optimizer and data augmentation approaches, achieving considerably better performance than ViT.  The following works including TNT~\cite{han2021transformer} and T2T~\cite{yuan2021tokens} enhance the effectiveness of modeling local structure, achieving better performance.  In parallel, PVT~\cite{wang2021pyramid} and PiT~\cite{heo2021pit} adopt a pyramid structure following CNNs, also attaining higher accuracy than DeiT and ViT. CPVT~\cite{chu2021conditional} improves the positional encoding. CaiT~\cite{touvron2021going} investigates deeper structures. Swin~\cite{liu2021swin} exploits the locality, achieving higher efficiency.  Container~\cite{gao2021container} proposes a hybrid architecture that exploits both convolution and Transformer, achieving excellent accuracy and efficiency.
Compared with the CNN-based methods,  methods based on vision Transformer have achieved comparable or even better performance. Besides, Transformer-based methods need less inductive bias, which is more friendly to network architecture search (NAS).

At last, we compare with MLP-based backbones including  FF~\cite{melas2021you}, MLP-mixer~\cite{tolstikhin2021mlp}, ResMLP~\cite{touvron2021resmlp} and S$^2$-MLP~\cite{yu2021s}. To make a fair comparison, we re-implement  both MLP-mixer~\cite{tolstikhin2021mlp} and ResMLP~\cite{touvron2021resmlp} through the same settings as ours and DeiT. As shown in Table~\ref{main}, using the same settings as ours and DeiT, Mixer-B/16$^*$  and ResMLP-36$^*$ achieve better performance than that in the original papers, Mixer-B/16~\cite{tolstikhin2021mlp} and ResMLP-36~\cite{touvron2021resmlp}.   By replacing the vanilla token-mixing MLP with the proposed CCS token-mixing MLP, both MLP-mixer and ResMLP achieve higher recognition accuracy with fewer parameters.  To be specific, using the vanilla token-mixing MLP, Mixer-B/16$^*$ only achieves a $77.2\%$ top-1 accuracy with $59$M parameters. In contrast, using our CCS token-mixing MLP, we achieve a $79.8\%$ 
 top-1 accuracy with only $57$M parameters. Besides, ResMLP-36$^*$ achieves a $79.8\%$ top-1 accuracy with $44$M parameters using the vanilla token-mixing MLP.  After using our CCS token-mixing MLP, the  top-1 accuracy increases to $80.6\%$ with only $43$M parameters. In fact, since the number of patches, $S=196$, is not large, using FFT for achieving the multiplication between a vector and a circulant matrix does not bring a considerable reduction in computation cost compared with the vanilla token-mixing MLP.  Meanwhile, compared with S$^2$-MLP-wide and S$^2$-MLP-deep,  Mixer-B/16$^*$+CCS  and ResMLP-36$^*$+CCS achieve comparable recognition accuracy  using fewer parameters and FLOPs. 
 

\subsection{Ablation study}
The core hyper-parameter in our CCS token-mixing MLP layer is $G$, the number of groups.  
We show the influence of $G$ on  ResMLP-36+CCS in Table~\ref{infG}.  As shown in Table~\ref{infG}, when $G=1$, it achieves a $80.0$, which has outperformed the original ResMLP-36$^{*}$~\cite{touvron2021resmlp}, validating the effectiveness of using circulant structure only.  Meanwhile, when $G$ increases from $1$ to $8$, the recognition on ImageNet1K consistently improves. This validates the effectiveness of conducting different token-mixing operations on different groups of channels. Besides, when $G$ further increases from $8$ to $384$, the recognition accuracy saturates but the number of parameters increases from $43$M to $46$M. Considering both effectiveness and efficiency, we set the number of groups, $G=8$, by default.


\begin{table}[htp!]
\centering
\begin{tabular}{c|cccc|c}
\hlineB{3}
G     &  1   & 4   & 8  & 384  & ResMLP-36$^{*}$~\cite{touvron2021resmlp} \\ \hline
top-1    &   $80.0$  &   $80.4$  & $\mathbf{80.6}$   &    $\mathbf{80.6}$  & $79.8$\\ 
top-5         &  $95.0$    &  $95.1$    &  $\mathbf{95.3}$  & $\mathbf{95.3}$ & $94.7$    \\
parameters   & $43$M    & $43$M    &  $43$M  &  $46$M  & $44$M \\ \hlineB{3}
\end{tabular}
\vspace{0.1in}
\caption{The influence of the number of groups, $G$, on the recognition accuracy and the number of parameters. The experiments are conducted on ImageNet1K dataset. }
\label{infG}
\end{table}

\section{Conclusion}
The token-mixing MLP in the existing MLP-based vision backbone is spatial-specific and channel-agnostic. The spatial-specific configuration makes it sensitive to spatial translation. Meanwhile, the channel-agnostic property limits its capability in mixing tokens. To overcome those limitations, we propose a Circulant Channel-specific (CCS) token-mixing MLP, which is spatial-agnostic and channel-specific. The spatial-agnostic property makes our CCS token-mixing MLP more robust to spatial translation and the channel-specific property enables it to encode richer visual patterns. Meanwhile, by exploiting the circulant structure, the number of parameters in the token-mixing MLP layer is significantly reduced.  Experiments on ImageNet1K dataset show that, by replacing existing token-mixing MLPs with our CCS token-mixing MLPs,  we achieve higher recognition accuracy with fewer parameters. 

\vspace{1in}

\bibliographystyle{plain}
\bibliography{cv}

\end{document}